\documentclass[10pt,twocolumn,letterpaper]{article}

\usepackage{iccv}
\usepackage{times}
\usepackage{epsfig}
\usepackage{graphicx}
\usepackage{amsmath}
\usepackage{amssymb}
\usepackage{mathtools}
\usepackage{multirow}
\usepackage{enumitem}
\usepackage{caption}
\usepackage{subcaption}
\usepackage{array}
\usepackage{colortbl}
\usepackage{booktabs}
\usepackage{bbm}
\usepackage{multicol}
\usepackage{makecell}
\usepackage{xcolor}
\usepackage{xspace}
\usepackage{float}
\usepackage{color}
\usepackage{pifont}
\setitemize[1]{itemsep=0pt,partopsep=0pt,parsep=\parskip,topsep=5pt}

\usepackage[pagebackref=true,breaklinks=true,letterpaper=true,colorlinks,bookmarks=false]{hyperref}

\iccvfinalcopy

\def\iccvPaperID{****}

\ificcvfinal\fi

\newlength\savedwidth

\definecolor{mygray}{gray}{.92}

\RequirePackage{geometry}
\AtBeginDocument{
  \baselineskip 14pt plus .2pt minus .2pt
}

\setlist{nosep}  
\setlist[enumerate,1]{label=(\arabic*)}  

\usepackage{titlesec}
\usepackage{xcolor}
\definecolor{main}{HTML}{7A2A87}

\titleformat{\section}
  {\color{main}\bfseries\fontsize{12pt}{14pt}\selectfont} 
  {\thesection} 
  {1em} 
  {} 

\titleformat{\subsection}
  {\bfseries\fontsize{10.5pt}{13.5pt}\selectfont}
  {\thesubsection}
  {.8em}
  {}

\titleformat{\subsubsection}
  {\fontsize{10.5pt}{10.5pt}\selectfont \itshape}
  {\thesubsubsection}
  {.8em}
  {}

\titlespacing*{\section}{0pt}{1em plus .3ex minus .2ex}{0.7em plus .2ex}
\titlespacing*{\subsection}{0pt}{1em plus 1ex minus .2ex}{0.5em plus .2ex}
\titlespacing*{\subsubsection}{0pt}{0.6em plus 1ex minus .2ex}{0.3em plus .2ex}

\RequirePackage[expansion=true,protrusion=false]{microtype}

\sfcode`.=1200
\sfcode`:=1200

\usepackage[numbers,sort&compress]{natbib} 
\usepackage{overpic}  

\newcommand{\secref}[1]{Sec. \ref{#1}}
\newcommand{\figref}[1]{Fig. \ref{#1}}
\newcommand{\tabref}[1]{Tab. \ref{#1}}
\newcommand{\eqnref}[1]{Eq. (\ref{#1})}

\newcommand{\pOne}{{PraNet-V1}}
\newcommand{\pTwo}{{PraNet-V2}}
\newcommand{\sArt}{{state-of-the-art~}}

\makeatletter
\renewcommand{\@makefnmark}{\hbox{\textsuperscript{\normalfont\@thefnmark}}} 
\makeatother

\begin{document}
	
	\title{\pTwo: Dual-Supervised Reverse Attention for Medical Image Segmentation}
	
	\author{
		Bo-Cheng Hu$^{1,2}$, 
		Ge-Peng Ji$^{3}$,
		Dian Shao$^{4}$,
		Deng-Ping Fan$^{1,2}$,\\
		$^1$Nankai Institute of Advanced Research (SHENZHEN-FUTIAN)~~~\\
		$^2$VCIP \& CS, Nankai University~~~
		$^3$School of Computing, Australian National University~~~\\
		$^4$Unmanned System  Research Institute, Northwestern Polytechnical University~~~
	}
	
	\maketitle
	\ificcvfinal\thispagestyle{empty}\fi

	\begin{abstract}
		Accurate medical image segmentation is essential for effective diagnosis and treatment. Previously, \pOne~was proposed to enhance polyp segmentation by introducing a reverse attention (RA) module that utilizes background information. However, \pOne~struggles with multi-class segmentation tasks. To address this limitation, we propose \pTwo, which, compared to \pOne, effectively performs a broader range of tasks including multi-class segmentation. At the core of \pTwo~is the Dual-Supervised Reverse Attention (DSRA) module, which incorporates explicit background supervision, independent background modeling, and semantically enriched attention fusion. Our \pTwo~framework demonstrates strong performance on four polyp segmentation datasets. Additionally, by integrating DSRA to iteratively enhance foreground segmentation results in three \sArt semantic segmentation models, we achieve up to a \textbf{1.36\%} improvement in mean Dice score. Code is available at: \url{https://github.com/ai4colonoscopy/PraNet-V2/tree/main/binary_seg/jittor}. 
	\end{abstract}

\section{Introduction}

\begin{figure}[ht]
\centering
\setlength{\fboxrule}{0pt}
\fbox{\includegraphics[width=0.47\textwidth]{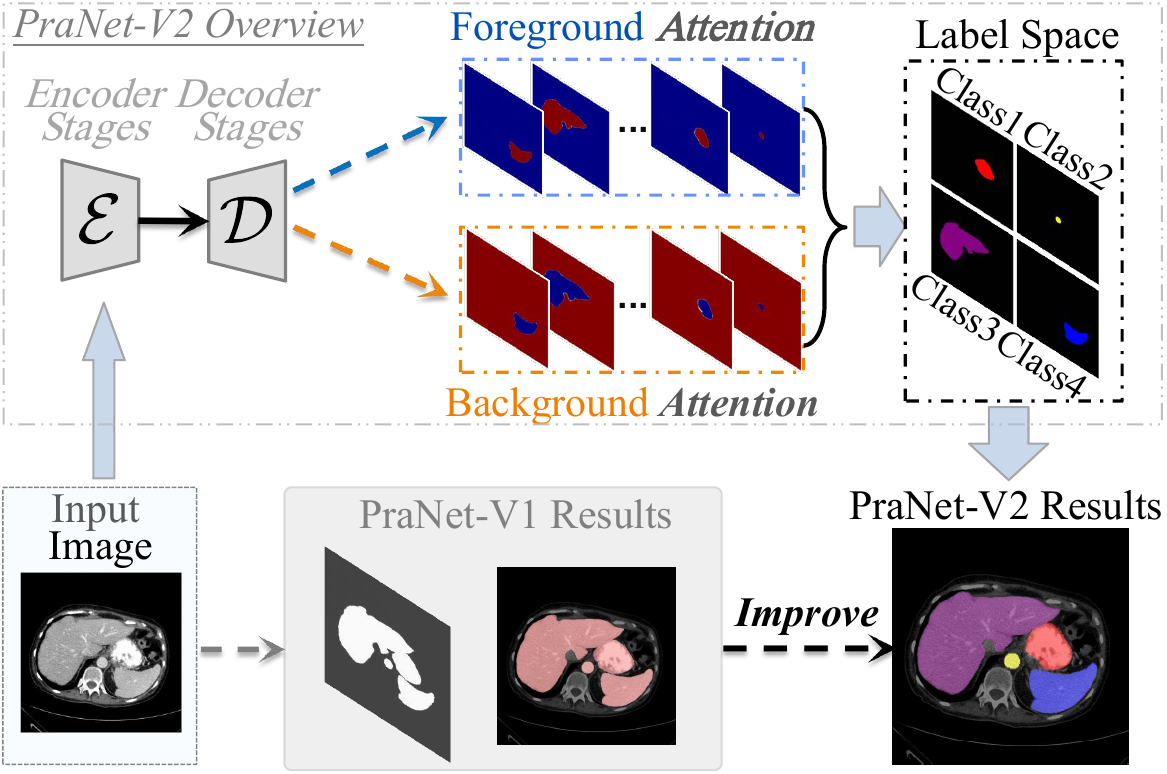}}
\caption{Illustration of the key differences between \pOne~and \pTwo~in background modeling and task handling.}
\label{fig:teaser}
\end{figure}

Medical image segmentation plays a vital role in modern medical diagnosis and treatment by identifying regions of interest—such as lesions, organs, or tissues—from medical images. As modern medical science increasingly relies on imaging technology, segmentation tasks in medical images
have progressed from binary classification \cite{fan2020pranet,guo2020learn,dong2023polyppvt,shao2024Polyper,Li_ASPS_MICCAI2024,Wan_LSSNet_MICCAI2024} to more complex multi-class segmentation \cite{chen2021transunet,oneprompt2024wu,Wan_LKMUNet_MICCAI2024,Zhu_SelfRegUNet_MICCAI2024}. 
For example, the U-Net series \cite{ronneberger2015u,zhou2019unet++,huang2020unet3+,cciccek20163d} utilizes an encoder-decoder architecture with various skip connections to capture multi-scale information, providing refined semantic and spatial features for segmentation. Building upon various U-Net models, the nnU-Net \cite{isensee2021nnu} emphasizes versatility over architectural innovation. In contrast, the DeepLab series \cite{chen2017deeplab} expands receptive fields through dilated convolutions but is still constrained by CNNs' limitations in holistic modeling. Furthermore, Transformer-based models \cite{chen2021transunet,cao2021swin,huang2021missformer,wang2022mixed} improve segmentation by combining local features with global context, effectively managing long-range dependencies.

\begin{figure*}[t!]
    \centering
    \setlength{\fboxrule}{0pt}
    \fbox{\includegraphics[width=0.9\textwidth]{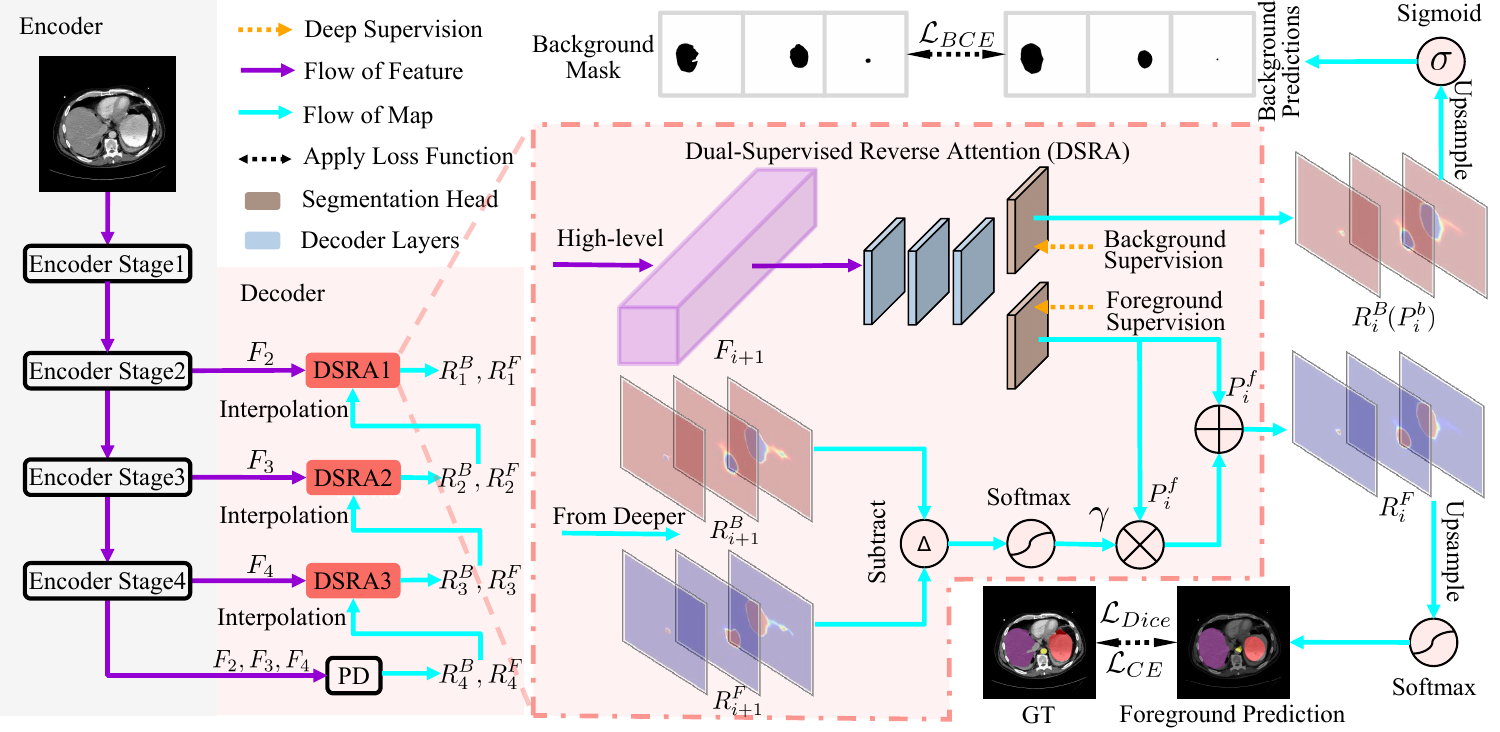}}
    \caption{Overview of the \pTwo~framework and DSRA module. The pipeline processes high-level features ($F_2,F_3,F_4$) through a parallel partial decoder (PD) and three DSRA modules. The DSRA module decodes \textcolor[rgb]{0.531, 0.102, 0.797}{high-level} feature $F_{i+1}$ to generate foreground and background segmentation maps, while integrating outputs \textcolor[rgb]{0, 1, 1}{from deeper} DSRA module or PD ($R_{i+1}^F$,$R_{i+1}^B$) to refine the foreground segmentation maps.}
    \label{fig:pipeline}
\end{figure*}

Despite advances in feature extraction and attention mechanisms, these models still \textit{neglect background features}. This oversight limits their ability to define boundaries accurately and diminishes performance in scenarios with low contrast between the foreground and background.
Our previous work, \pOne~\cite{fan2020pranet}, introduces reverse attention (RA) to explicitly model the background, enabling effective polyp segmentation under minimal contrast and significant class imbalance. While \pOne~pioneers reverse attention in medical image segmentation, further evaluations reveal several constraints: 
(1) \textbf{Limited application scenarios.} 
\pOne~is tailored for binary polyp segmentation, making it inadequate for multi-class segmentation; 
(2) \textbf{Rule-based direct inversion.} 
In \pOne, reverse attention weights are generated by directly subtracting each pixel’s foreground probability from one. It does not provide additional contextual information and inherits inaccuracies from the forward attention; 
(3) \textbf{Semantically ambiguous attention fusion.} 
\pOne~combines reverse and forward attention in the feature space, resulting in intertwined high-dimensional foreground and background features without clear semantic boundaries.
We tackle the challenges by retuning the RA module for multi-class segmentation. Illustrated in \figref{fig:teaser}, we introduce the Dual-Supervised Reverse Attention (DSRA) module, which uses individual parameters to explicitly learn foreground-background attention of each class. 
Additionally, the DSRA module fuses foreground and background information within a semantically enriched label space, refining interpretability compared to the traditional RA module. 
Overall, the main contributions of our work are as follows:
\begin{enumerate}
    \item \textbf{Novel module:} 
    We introduce the DSRA module, which separates foreground and background processing to enhance background and edge recognition.

    \item \textbf{New framework:} 
    Utilizing DSRA, we develop the \pTwo~framework, which outperforms \pOne~in the polyp segmentation task.


    \item \textbf{New SOTA results:} 
    By integrating DSRA into three leading segmentation models, we refine their segmentation outcomes, achieving mean Dice score improvements of 0.50\% to 1.36\% in mean Dice scores.
\end{enumerate}

\section{Method}

\subsection{Dual Supervised Reverse Attention Module}
\label{sec:DSRA}
In polyp segmentation, \pOne~effectively employs reverse attention (RA) for binary tasks by capturing background information. However, for multi-class segmentation, a single RA calculation fails to differentiate between classes, and its rule-based RA extraction is incompatible with multi-channel pixel-level confidence outputs.


Therefore, we introduce the Dual-Supervised Reverse Attention (DSRA) module, building upon the RA module from \pOne. 
As illustrated in \figref{fig:pipeline}, we organize DSRA modules as decoder stages in a U-Net structure, leveraging the advantages of multi-scale features and skip connections. 
For an input image $X \in \mathbb{R}^{H \times W \times C}$, the encoder generates four multi-scale features $\{F_i \,|\, i = 1, 2, 3, 4\}$, where $ F_i \in \mathbb{R}^{\frac{H}{2^{i+1}} \times \frac{W}{2^{i+1}} \times C_i}$ means the feature at the $i$-th encoder stage. The decoder then processes three high-level features $\{F_i \,|\, i=2,3,4\}$ to generate four segmentation outputs $\{R_i \,|\, i=1,2,3,4\}$ across four stages, comprising a parallel partial decoder (PD) \cite{fan2020pranet} followed by DSRA3 to DSRA1. Each segmentation output includes results for each semantic class ($R_i^F$) and their corresponding background regions ($R_i^B$). 
We modify the output layers of the PD and use it in the first decoder stage to aggregate high-level features ($F_2$, $F_3$, $F_4$), providing the coarse segmentation result $R_4 = \{R_4^F, R_4^B\}$.

Note that the subsequent three stages adopt DSRA modules to transform coarse predictions into fine results. As shown on the right side of \figref{fig:pipeline}, the output of the $i$-th DSRA module ($i=1,2,3$), $R_i = \{R_i^F, R_i^B\}$, is computed as follows:
\begin{align}
    R_{i}^{F} & = P^{f}_{i}+ P^{f}_{i}\circ \gamma, \label{eq:1}\\
    R_{i}^{B} & = P^{b}_{i}.\label{eq:2}
\end{align}
Here, $P^{f}_{i}$ and $P^{b}_{i}$ denote the outputs of the segmentation heads under foreground and background supervision, respectively. The symbol $\circ$ represents element-wise multiplication. The term $\gamma$, referred to as the \textit{reverse gain}, incorporates refinement information from a deeper DSRA module to better leverage the cascade structure and background information. 
Unlike the RA module in \pOne, the DSRA module employs dedicated supervision and structures to \textbf{separately} produce foreground and background segmentation results, avoiding the entanglement seen in shared structures and parameters. The calculations for $P_i^f$, $P_i^b$, and $\gamma$ are as follows:
\begin{flalign}
&\ \{P^{f}_{i},P_{i}^{b}\} = \phi(F_{i+1}), \label{eq:3}\\
&\ \gamma=\text{Softmax}( \mathcal{I}(R_{i+1}^{F};F_{i+1}) -  \mathcal{I}(R_{i+1}^{B};F_{i+1})),
\label{eq:4}
\end{flalign}
where $\phi(\cdot)$ denotes convolutional layers forming the decoder layers and segmentation heads. $\mathcal{I}(x;y)$ resizes the width and height of $x$ to match $y$ using bilinear interpolation. $R_{i+1}^{F}$ and $R_{i+1}^{B}$ respectively represent the foreground and background segmentation maps from the DSRA module in layer $i+1$. In the hierarchical cascade described by the above formulas, outputs from deeper DSRA modules are progressively integrated to refine coarse segmentation results. 
These modifications enable DSRA modules to use parameter-based learning to extract reverse attention and fuse it with forward attention in the segmentation label space. This method resolves the spatial and semantic misalignment issues inherent in \pOne's direct operation on compressed decoder features.
In summary, DSRA excels over RA in three key aspects:

(1) \textbf{Independent Structure}: DSRA utilizes \textbf{separate} segmentation heads for foreground and background, enhancing feature detail capture.

(2) \textbf{Background Modeling}: DSRA adopts additional supervision to learn objects’ background for each class via parameter fitting, enabling effective multi-class segmentation.

(3) \textbf{Semantic Information Refinement}: DSRA fuses foreground and background information in the label space, fully leveraging pixel-level confidence to enhance boundary and background accuracy.

\subsection{Background Supervision and Loss Function}
\noindent\textbf{Background Mask. }
To supervise each semantic class's background, we introduce a multi-channel background mask, where each channel corresponds to an object class. In this mask, a pixel value of 1 indicates background regions, and 0 represents the object. As illustrated by the `Background Mask' at the top of \figref{fig:pipeline}, it extracts the background for each semantic class from the ground truth, where white denotes 1 and black denotes 0.

\noindent\textbf{Loss Function. }
Leveraging the background mask, the total loss is defined as $\mathcal{L}_{\text{total}} = w_1 \times \mathcal{L}_{\text{Dice}} + w_2 \times \mathcal{L}_{\text{CE}} + w_3 \times \mathcal{L}_{\text{BCE}}$, where $w_1$, $w_2$, and $w_3$ are weighting coefficients. For foreground supervision, the Dice loss ($\mathcal{L}_{\text{Dice}}$) addresses class imbalance and ensures region overlap, while the cross-entropy loss ($\mathcal{L}_{\text{CE}}$) enhances pixel-level classification. For background supervision, the binary cross-entropy loss ($\mathcal{L}_{\text{BCE}}$) aligns background predictions with the background mask, enabling independent background learning for each class.

\noindent\textbf{Implementation. }
Building on DSRA, we propose \pTwo~framework, with its polyp segmentation performance detailed in \secref{sec:bcs}. In addition, the DSRA demonstrates high versatility. Most mainstream segmentation networks can utilize two segmentation heads, like DSRA, to separately generate foreground and background segmentation results, and then iteratively enhance foreground segmentation as described in \eqnref{eq:1} and \eqnref{eq:4}. Leveraging this flexibility, we further evaluate DSRA's integration into three state-of-the-art models for multi-class medical image segmentation in  \secref{sec:mcs}.

\section{Experiment}

\subsection{Binary Class Segmentation}
\label{sec:bcs}

\begin{table*}[t]
\centering
\small
\caption{Performance comparison of \pOne~and \pTwo~on four polyp segmentation datasets, with best values in bold.}
\setlength\tabcolsep{4.8pt}
\renewcommand{\arraystretch}{1.1}
\begin{tabular}{rccccccccccccp{10mm}}
\hline
\multirow{2}{*}{Dataset} & \multirow{2}{*}{Backbone} & \multicolumn{2}{c}{mDice (\%)} & \multicolumn{2}{c}{mIoU (\%)} & \multicolumn{2}{c}{wFm (\%)} & \multicolumn{2}{c}{S-m (\%)} & \multicolumn{2}{c}{mEm (\%)} & \multicolumn{2}{c}{MAE ($\times 10^{-2}$)} \\ \cline{3-14} 
 &  & V1 & V2 & V1 & V2 & V1 & V2 & V1 & V2 & V1 & V2 & V1 & V2 \\ \hline
CVC-300 & \multirow{4}{*}{Res2Net50 \cite{gao2021res2net}} & 87.06 & \textbf{89.83} & 79.61 & \textbf{82.66} & 84.32 & \textbf{87.79} & 92.55 & \textbf{93.70} & 94.97 & \textbf{97.47} & 0.99 & \textbf{0.59} \\
CVC-ClinicDB &  & 89.84 & \textbf{92.28} & 84.83 & \textbf{87.22} & 89.63 & \textbf{91.97} & 93.67 & \textbf{94.87} & 96.22 & \textbf{97.38} & 0.94 & \textbf{0.91} \\
Kvasir &  & 89.39 & \textbf{90.70} & 83.55 & \textbf{85.29} & 88.00 & \textbf{89.59} & 91.25 & \textbf{91.70} & 94.00 & \textbf{95.07} & 3.04 & \textbf{2.35} \\
ETIS &  & 62.75 & \textbf{64.05} & \textbf{56.57} & 56.54 & 60.07 & \textbf{60.43} & 79.33 & \textbf{79.41} & \textbf{80.77} & 79.74 & 3.07 & \textbf{2.08} \\ \hline
CVC-300 & \multicolumn{1}{l}{\multirow{4}{*}{PVTv2-B2 \cite{wang2022pvt}}} & 86.59 & \textbf{89.89} & 78.92 & \textbf{83.11} & 83.15 & \textbf{88.48} & 91.84 & \textbf{93.96} & 94.45 & \textbf{97.04} & 1.03 & \textbf{0.73} \\
CVC-ClinicDB & \multicolumn{1}{l}{} & 90.96 & \textbf{93.09} & 85.42 & \textbf{88.06} & 89.90 & \textbf{92.80} & 94.34 & \textbf{94.45} & 96.49 & \textbf{98.23} & 1.02 & \textbf{0.84} \\
Kvasir & \multicolumn{1}{l}{} & 87.09 & \textbf{91.52} & 81.31 & \textbf{86.12} & 84.52 & \textbf{90.39} & 89.33 & \textbf{92.50} & 92.58 & \textbf{95.64} & 4.19 & \textbf{2.33} \\
ETIS & \multicolumn{1}{l}{} & 68.32 & \textbf{76.35} & 60.02 & \textbf{68.72} & 61.65 & \textbf{72.96} & 81.38 & \textbf{86.50} & 80.92 & \textbf{88.26} & 4.14 & \textbf{1.45} \\ \hline
\end{tabular}
\label{tab:binary_exp}
\end{table*}


\textbf{Datasets. }
We conduct experiments on four polyp segmentation datasets to compare the performance of \pOne~and \pTwo, following the data split methodology of \pOne~\cite{fan2020pranet}. Specifically, the datasets include CVC-ClinicDB \cite{BERNAL201599}, CVC-300 \cite{vazquez2017benchmark}, Kvasir \cite{jha2020kvasir}, and ETIS \cite{Silva_2013}. 
The CVC-ClinicDB dataset contains 612 images, of which 62 are used for testing and the rest for training. The ETIS dataset includes 196 challenging images, containing polyps that are typically small and
difficult to detect. The Kvasir dataset contains 1,000 images with 700 large, 48 small, and 323 medium polyps \cite{kim2021uacanet}, and is randomly divided into 80\% training, 10\% validation, and 10\% testing. For generalization evaluation, the ETIS and CVC-300 datasets are used exclusively, with their images omitted from training.

\noindent\textbf{Training Details. }
The experiments are performed on a distributed cluster with an NVIDIA GeForce RTX 3090 GPU, using PyTorch 2.0.1 and CUDA 12.2. The training protocols are consistent with those of \pOne, including resizing inputs to $352 \times 352$ and employing a multi-scale training strategy \{0.75$\times$, 1$\times$, 1.25$\times$\}. The Adam optimizer is applied with a fixed learning rate of $1 \times 10^{-4}$. We extend the loss function of \pOne~by adding $\mathcal{L}_{\text{BCE}}^{w}$ to supervise one DSRA segmentation head, applying pixel-wise weighting to penalize boundary errors more heavily and sharpen the edges.

\noindent\textbf{Evaluation Metrics. }
Following \pOne~\cite{fan2020pranet}, we evaluate segmentation performance using metrics including mean Dice (mDice, \%), mean IoU (mIoU, \%), weighted F-measure (wFm, \%), structure measure (S-m, \%) \cite{fan2017structure}, mean enhanced measure (mEm, \%) \cite{Fan2018Enhanced}, and mean absolute error (MAE). Among these, mDice, mIoU, and MAE are classic binary segmentation metrics, while wFm evaluates segmentation by assigning greater weight to polyp interiors and edges. Beyond pixel-wise metrics, S-m evaluates segmentation by jointly considering region-based and object-based structural similarity, whereas mEm integrates pixel-level alignment with global structural correctness for a comprehensive assessment.

\noindent\textbf{Quantitative Results. } 
To ensure a fair comparison, we implement \pOne~and \pTwo~with the same backbone. As shown in \tabref{tab:binary_exp}, when using Res2Net50 \cite{gao2021res2net} as the backbone, \pTwo~consistently outperforms \pOne~on almost all datasets and metrics. Specifically, on the CVC-300 and CVC-ClinicDB datasets, \pTwo~achieves mDice improvements of 2.77\% and 2.44\%, and mIoU gains of 3.05\% and 2.39\%, respectively, demonstrating enhanced segmentation accuracy. 
When employing PVTv2-B2 \cite{wang2022pvt} as the backbone, \pTwo~shows even more pronounced performance gains. It consistently surpasses \pOne~on all benchmark datasets and metrics, underscoring DSRA's robustness across diverse encoder architectures. On the \textit{unseen} ETIS dataset, \pTwo~exhibits a remarkable 8.03\% increase in mDice and an impressive 8.70\% boost in mIoU, demonstrating its strong generalization ability. Furthermore, \pTwo~shows a 5.12\% improvement in S-m, indicating our model preserves the overall shape, boundary consistency, and geometric integrity of the segmented regions. 


\subsection{Multi-Class Segmentation}
\label{sec:mcs}

\begin{table*}[t]
\centering
\small
\caption{Performance comparison on the Synapse dataset. Baseline results are sourced from \cite{rahman2024emcad} and values improved by our DSRA are in bold. \textsuperscript{\textdagger} indicates reproduced performance. Dice scores (\%) for each organ are also included.}
\label{tab:inte_synapse}
\renewcommand\arraystretch{0.9}
\setlength\tabcolsep{5.1pt}
\begin{tabular}{rccccccccccc}
\hline
Architectures & mDice (\%) & HD95 (mm) & mIoU (\%) & Aorta & GB & KL & KR & Liver & PC & SP & SM \\
\hline
UNet \cite{ronneberger2015u} & 70.11 & 44.69 & 59.39 & 84.00 & 56.70 & 72.41 & 62.64 & 86.98 & 48.73 & 81.48 & 67.96 \\
AttnUNet \cite{oktay2018attention} & 71.70 & 34.47 & 61.38 & 82.61 & 61.94 & 76.07 & 70.42 & 87.54 & 46.70 & 80.67 & 67.66 \\
R50+UNet \cite{chen2021transunet} & 74.68 & 36.87 & -- & 84.18 & 62.84 & 79.19 & 71.29 & 93.35 & 48.23 & 84.41 & 73.92 \\
R50+AttnUNet \cite{chen2021transunet} & 75.57 & 36.97 & -- & 55.92 & 63.91 & 79.20 & 72.71 & 93.56 & 49.37 & 87.19 & 74.95 \\
SSFormer \cite{wang2022stepwise} & 78.01 & 25.72 & 67.23 & 82.78 & 63.74 & 80.72 & 78.11 & 93.53 & 61.53 & 87.07 & 76.61 \\
PolypPVT \cite{dong2023polyppvt} & 78.08 & 25.61 & 67.43 & 82.34 & 66.14 & 81.21 & 73.78 & 94.37 & 59.34 & 88.05 & 79.40 \\
TransUNet \cite{chen2021transunet} & 77.61 & 26.90 & 67.32 & 86.56 & 60.43 & 80.54 & 78.53 & 94.33 & 58.47 & 87.06 & 75.00 \\
SwinUNet \cite{cao2021swin} & 77.58 & 27.32 & 66.88 & 81.76 & 65.95 & 82.32 & 79.22 & 93.73 & 53.81 & 88.04 & 75.79 \\
MT-UNet \cite{wang2022mixed} & 78.59 & 26.59 & -- & 87.92 & 64.99 & 81.47 & 77.29 & 93.06 & 59.46 & 87.75 & 76.81 \\
MISSFormer \cite{huang2021missformer} & 81.96 & 18.20 & -- & 86.99 & 68.65 & 85.21 & 82.00 & 94.41 & 65.67 & 91.92 & 80.81 \\
PVT-CASCADE \cite{Rahman_2023_WACV} & 81.06 & 20.23 & 70.88 & 83.01 & 70.59 & 82.23 & 80.37 & 94.08 & 64.43 & 90.10 & 83.69 \\
TransCASCADE \cite{Rahman_2023_WACV} & 82.68 & 17.34 & 73.48 & 86.63 & 68.48 & 87.66 & 84.56 & 94.43 & 65.33 & 90.79 & 83.52 \\
\hline
MIST\textsuperscript{\textdagger} \cite{rahman2023mist} & 81.91 & 14.93 & -- & 86.15 & 71.43 & 83.09 & 76.43 & 96.02 & 68.20 & 89.39 & 84.59 \\
\textbf{MIST \textit{(w/ DSRA)}} & \textbf{83.27} & \textbf{14.11} & -- & \textbf{87.54} & \textbf{75.36} & 82.23 & \textbf{76.53} & 95.93 & \textbf{71.51} & \textbf{91.66} & \textbf{85.41} \\
\hline
EMCAD-B2\textsuperscript{\textdagger} \cite{rahman2024emcad} & 82.71 & 21.74 & 74.65 & 87.24 & 69.56 & 85.23 & 80.88 & 95.59 & 65.88 & 92.62 & 84.64 \\
\textbf{EMCAD-B2 \textit{(w/ DSRA)}} & \textbf{83.75} & \textbf{17.77} & \textbf{74.81} & \textbf{88.69} & \textbf{72.79} & \textbf{85.41} & \textbf{82.91} & \textbf{95.82} & \textbf{68.47} & \textbf{93.09} & \textbf{85.85} \\
\hline
\end{tabular}
\end{table*}

\begin{table}[ht]
\centering
\small
\caption{Comparison on the ACDC dataset, with Dice scores (\%) for each class. Baseline results are from \cite{rahman2023multi,rahman2023mist} and values improved by our DSRA are in bold. \textsuperscript{\textdagger} denotes reproduced performance.}
\label{tab:inte_ACDC}
\setlength{\tabcolsep}{2.5pt} 
\renewcommand{\arraystretch}{0.95}
\begin{tabular}{rcccc}
\hline
Architectures & mDice (\%) & RV & Myo & LV \\
\hline
R50+UNet \cite{chen2021transunet} & 87.55 & 87.10 & 80.63 & 94.92 \\
R50+AttnUNet \cite{chen2021transunet} & 86.75 & 87.58 & 79.20 & 93.47 \\
ViT+CUP \cite{chen2021transunet} & 81.45 & 81.46 & 70.71 & 92.18 \\
R50+ViT+CUP \cite{chen2021transunet} & 87.57 & 86.07 & 81.88 & 94.75 \\
TransUNet \cite{chen2021transunet} & 89.71 & 88.86 & 84.53 & 95.73 \\
SwinUNet \cite{cao2021swin} & 90.00 & 88.55 & 85.62 & 95.83 \\
MT-UNet \cite{wang2022mixed} & 90.43 & 86.64 & 89.04 & 95.62 \\
MISSFormer \cite{huang2021missformer} & 90.86 & 89.55 & 88.04 & 94.99 \\
PVT-CASCADE \cite{Rahman_2023_WACV} & 91.46 & 88.90 & 89.97 & 95.50 \\
nnUNet \cite{isensee2021nnu} & 91.61 & 90.24 & 89.24 & 95.36 \\
nnFormer \cite{zhou2021nnformer} & 91.78 & 90.22 & 89.53 & 95.59 \\
\hline
MIST\textsuperscript{\textdagger} \cite{rahman2023mist} & 91.73 & 89.98 & 89.39 & 95.84 \\
\textbf{MIST \textit{(w/ DSRA)}} & \textbf{92.31} & \textbf{90.82} & \textbf{90.07} & \textbf{96.04} \\
\hline
Cascaded MERIT\textsuperscript{\textdagger} \cite{rahman2023multi} & 91.78 & 90.36 & 89.21 & 95.79 \\
\textbf{Cascaded MERIT \textit{(w/ DSRA)}} & \textbf{92.28} & \textbf{91.27} & \textbf{89.38} & \textbf{96.19} \\
\hline
\end{tabular}
\end{table}

\noindent\textbf{Models and Datasets. } 
We conduct experiments on two multi-class medical image segmentation datasets to evaluate the performance of \sArt models after integrating DSRA. The models used include Cascaded MERIT \cite{rahman2023multi}, MIST \cite{rahman2023mist}, and EMCAD \cite{rahman2024emcad}. For the datasets, we use the Automated Cardiac Diagnosis Challenge dataset (ACDC) \cite{ACDC} and the Synapse multi-organ segmentation dataset (Synapse) \cite{landman2015miccai}. The ACDC dataset contains cardiac MRI scans from 100 patients, annotated for three structures: right ventricle (RV), left ventricle (LV), and myocardium (Myo). The Synapse dataset comprises 30 contrast-enhanced CT scans (3,779 slices) with annotations for 8 abdominal organs, including the aorta (Aorta), gallbladder (GB), left kidney (KL), right kidney (KR), liver (Liver), pancreas (PC), spleen (SP), and stomach (SM). We follow the dataset splits used in the aforementioned three models to ensure evaluation consistency.



\noindent\textbf{Training Details. } 
The original training protocols of the three models are adopted, with minor batch size adjustments made to better accommodate the DSRA module.

\noindent\textbf{Evaluation Metrics. }
Following \cite{rahman2023multi,rahman2023mist,rahman2024emcad}, segmentation performance is evaluated using mean Dice (mDice, \%) on the ACDC and Synapse datasets. Additionally, the Synapse dataset incorporates mean IoU (mIoU, \%) and 95th Percentile Hausdorff Distance (HD95, mm) following \cite{rahman2024emcad}. HD95 measures the maximum deviation between predicted and ground-truth boundaries within the 95th percentile. Lower HD95 values indicate better boundary alignment.

\begin{figure*}[p]
\centering
    \begin{overpic}[width=.95\textwidth]{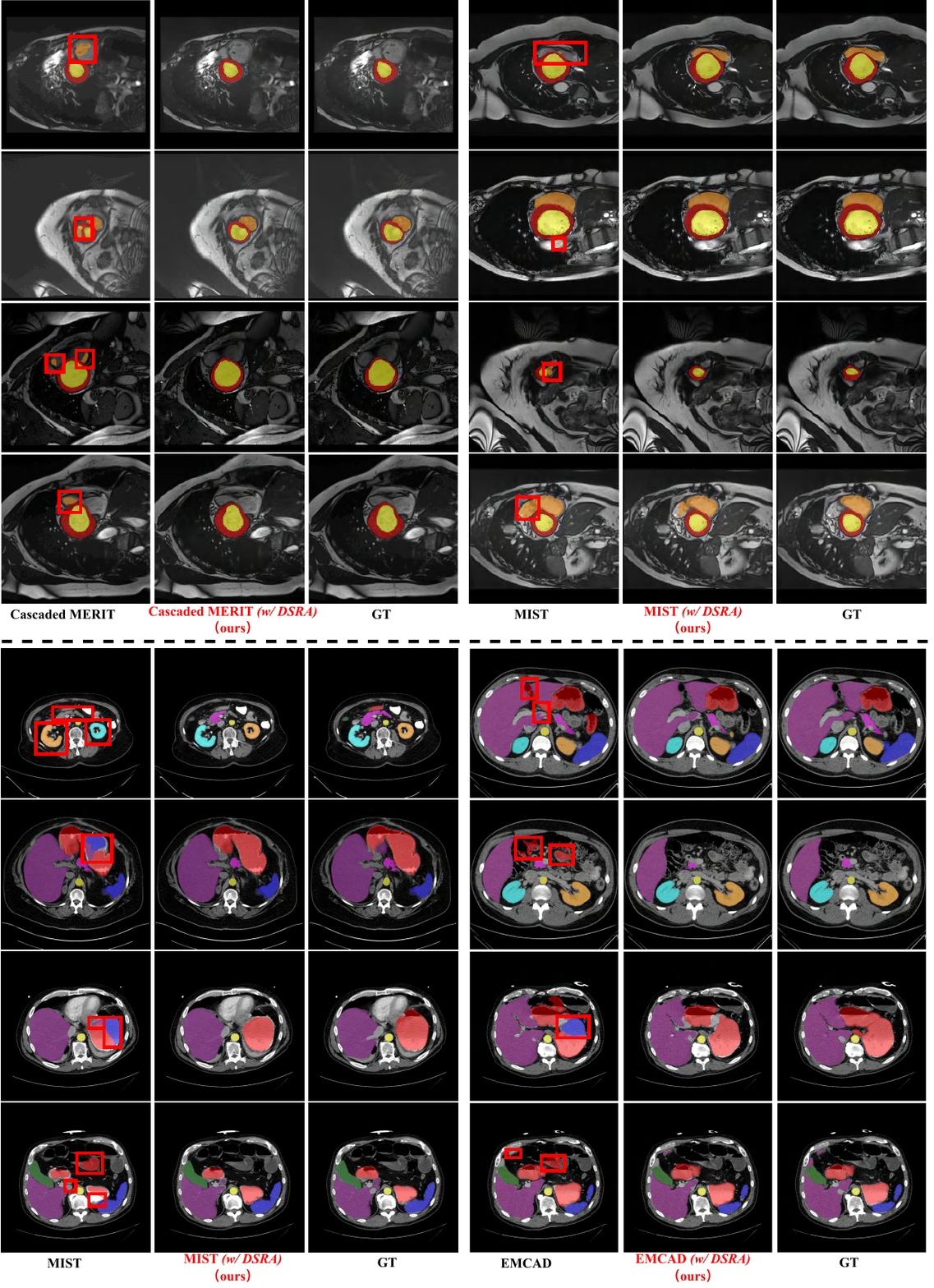}
    \put(9.3,51.8) {\footnotesize \cite{rahman2023multi}}
    \put(43,51.8) {\footnotesize \cite{rahman2023mist}}
    \put(6.7,1.3) {\footnotesize \cite{rahman2023mist}}
    \put(43.2,1.3) {\footnotesize \cite{rahman2024emcad}}
    \end{overpic}
    \caption{Segmentation results on ACDC (above) and Synapse (below) datasets, with segmentation errors highlighted in \textcolor{red}{red boxes}.}
    \label{fig:vis_all}
\end{figure*}

\noindent\textbf{Quantitative Results. }
On the Synapse dataset with eight categories of organs, the benefits of DSRA integration are evident (\tabref{tab:inte_synapse}). MIST (\textit{w/ DSRA}) achieves an average Dice score of 83.27\% and reduces HD95 to 14.11, surpassing its original version by 1.36\% and 0.82, respectively. This improvement is further reflected in organ-level performance, where the Dice score for the GB (gallbladder) increases significantly from 71.43\% to 75.36\% (+3.93\%). Similarly, EMCAD-B2 (\textit{w/ DSRA}) demonstrates robust performance gains, with the average Dice score increasing from 82.71\% to 83.75\% (+1.04\%) and HD95 dropping from 21.74 to 17.77. These results emphasize the DSRA module's ability to boost segmentation accuracy and reduce prediction errors.

On the ACDC dataset, as shown in \tabref{tab:inte_ACDC}, the integration of the DSRA module also enhances the performance of both MIST and Cascaded MERIT. MIST (\textit{w/ DSRA}) and Cascaded MERIT (\textit{w/ DSRA}) achieve average Dice scores of 92.31\% and 92.28\%, respectively, reflecting consistent advancements over their original versions. Notably, both models show nearly a 1\% improvement in the Dice score for the RV (right ventricle), a challenging class due to its irregular shape.

\noindent\textbf{Qualitative Results. }
\figref{fig:vis_all} visualizes segmentation results on the ACDC and Synapse datasets. It compares the performance of the three models before and after integrating the DSRA. We can see that with DSRA integration, all three models present more accurate results with fewer errors and redundancies.



\begin{table}[t]
\centering
\small
\caption{Ablation on loss function.}
\setlength{\tabcolsep}{11.2pt} 
\renewcommand\arraystretch{0.9} 
\begin{tabular}{ccccc}
\hline
\multicolumn{3}{c}{Loss function} & & \\
\cline{1-3} $\mathcal{L}_{\text{BCE}}$ & $\mathcal{L}_{\text{CE}}$ & $\mathcal{L}_{\text{Dice}}$ & \multirow{-2}{*}{mDice (\%)} & \multirow{-2}{*}{mIoU (\%)} \\
\hline
\ding{51} & \ding{51} & & 91.91 & 85.43 \\
\ding{51} & & \ding{51} & 92.14 & 85.72 \\
& \ding{51} & \ding{51} & 92.16 & 85.84 \\
\ding{51} & \ding{51} & \ding{51} & \textbf{92.31} & \textbf{86.02} \\
\hline
\end{tabular}

\label{tab:Ablation_loss}
\end{table}

\subsection{Ablation Study}
\label{sec:ab}

We conduct ablation experiments on the MIST \textit{(w/ DSRA)} model to assess the impact of different combinations of loss functions on segmentation performance. As summarized in \tabref{tab:Ablation_loss}, the network achieves optimal performance with the combined use of $\mathcal{L}_{\text{BCE}}$, $\mathcal{L}_{\text{CE}}$, and $\mathcal{L}_{\text{Dice}}$, achieving an mDice of 92.31\% and an mIoU of 86.02\%. Removing any of these loss functions results in a performance drop. Specifically, excluding the $\mathcal{L}_{\text{Dice}}$ impairs the network's ability to handle class-imbalanced input images, while removing the $\mathcal{L}_{\text{BCE}}$ or $\mathcal{L}_{\text{CE}}$ weakens the ability of the two segmentation heads in the DSRA module to map features to the label space.

\section{Discussion and Conclusion}
\noindent\textbf{Limitations and Future Work. }
Although integrating DSRA enhances the performance of three state-of-the-art segmentation models, these models are limited to a fixed number of classes predefined by the training dataset, restricting their ability to handle unknown categories in open-world scenarios. Inspired by how medical practitioners rely on expert consensus for unclassified diseases, future work could explore anomaly detection and incorporate approaches such as dual-decoder architectures with feature space manipulation to identify and adapt to unknown categories \cite{Sodano2024openworld}.


\noindent\textbf{Conclusion.} 
In this paper, we address the limitations in \pOne~by introducing the Dual-Supervised
Reverse Attention (DSRA) module, resulting in the improved \pTwo~framework. 
By decoupling foreground and background feature computations, DSRA enhances feature separation and significantly refines segmentation accuracy in the polyp segmentation task. Furthermore, the versatility of DSRA is demonstrated through integration into three \sArt segmentation models. This integration achieves mean Dice score improvements ranging from 0.50\% to 1.36\% across two benchmark datasets. Such results underscore the module's broad applicability and effectiveness. 
Our future work will involve extending the reverse attention to solve the text-guided semantic segmentation problem \cite{ji2024frontiers}.

\bibliographystyle{ieee_fullname}
\bibliography{main}


\end{document}